\documentclass[10pt,twocolumn,letterpaper]{article}

\usepackage{iccv}
\usepackage{times}
\usepackage{epsfig}
\usepackage{graphicx}
\usepackage{amsmath}
\usepackage{amssymb}

\usepackage{algorithm}
\usepackage{algorithmic}
\newtheorem{definition}{Definition}

\graphicspath{{figures/}}

\usepackage[pagebackref=true,breaklinks=true,letterpaper=true,colorlinks,bookmarks=false]{hyperref}

\iccvfinalcopy % 

\ificcvfinal\fi

\begin{document}

%%%%%%%%% TITLE
\title{Causal Analysis for Robust Interpretability of Neural Networks}

\author{Ola Ahmad\\
Thales Digital Solutions\\
Montreal, Canada\\
%{\tt\small ola.ahmad@thalesdigital.io}
\and
Nicolas Bereux\thanks{This work has been done when the authors were working at Thales Digital Solutions Inc.}\\
Université Paris Sud\\
Paris, France\\
\and
Loïc Baret\\
Thales Digital Solutions\\
Quebec, Canada\\
\and
Vahid Hashemi\\
AUDI\\
Ingolstadt, Germany\\
\and
Freddy Lecue\footnotemark[1]\\
J.P. Morgan, Chase \& Co\\
New York, USA\\
}

%\and

\maketitle
%\ificcvfinal\thispagestyle{empty}\fi

%%%%%%%%% ABSTRACT
\begin{abstract}
Interpreting the inner function of neural networks is crucial for the trustworthy development and deployment of these black-box models. Prior interpretability methods focus on correlation-based measures to attribute model decisions to individual examples. However, these measures are susceptible to noise and spurious correlations encoded in the model during the training phase (e.g., biased inputs, model overfitting, or misspecification). Moreover, this process has proven to result in noisy and unstable attributions that prevent any transparent understanding of the model's behavior. In this paper, we develop a robust interventional-based method grounded by causal analysis to capture cause-effect mechanisms in pre-trained neural networks and their relation to the prediction. Our novel approach relies on path interventions to infer the causal mechanisms within hidden layers and isolate relevant and necessary information (to model prediction), avoiding noisy ones. The result is task-specific causal explanatory graphs that can audit model behavior and express the actual causes underlying its performance. We apply our method to vision models trained on classification tasks. On image classification tasks, we provide extensive quantitative experiments to show that our approach can capture more stable and faithful explanations than standard attribution-based methods. Furthermore, the underlying causal graphs reveal the neural interactions in the model, making it a valuable tool in other applications (e.g., model repair).
\end{abstract}

%%%%%%%%% BODY TEXT
\section{Introduction}

Explainability and interpretability are crucial for deep neural networks (DNNs), which are disseminated in many applications, including vision and natural language processing. Despite their popularity, their opaque nature limits the adoption of these "black-box" models in domains requiring critical decisions without the ability to understand their behavior. Attempts to provide a transparent understanding of DNN systems have led to the development of many interpretability methods. Most of them focus on interpreting the function of DNNs through correlation-based measures, which attribute the model's decision to individual inputs \cite{vig_causal_2020}. The most popular ones are saliency (or feature attribution) methods \cite{IG, saliency, exbackprob, occlusion, gradshape, gradxinp, nlpJuraf16, 8579008}.
%
%Therefore, several methods have been recently proposed to extract human-interpretable concepts from trained models. The literature points out two broad directions of DNN explainability; behaviour methods \cite{behavior1, influfunc_2017} and structural analyses \cite{vig_causal_2020, geiger_causal_2021}. The former interprets model behaviour by its performance on controlled set (e.g., training data or evaluation set). Structural methods seek explanations from the internal structure of DNNs. The most common are attribution (or features importance) methods such as \cite{IG, saliency, exbackprob, occlusion, gradshape, gradxinp, nlpJuraf16}. 
%However, existing tools have their limitations, and numerous lack a rigorous methodology [13]
%
\par
\begin{figure*}[t]
   \centering
   \includegraphics[width=\linewidth]{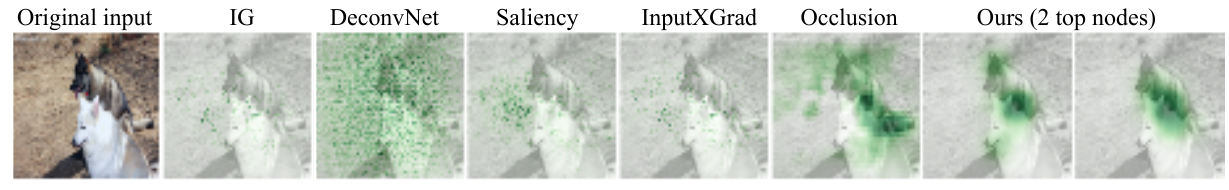}
   \caption{\textbf{Features importance from different explanation methods on a hard example from ImageNet data.} The actual class is "white wolf". The predicted class by the pre-trained ResNet18 is "Malamute". To the right is our method showing top 2 semantics (top head and body of wrong class) from the causal graph that explain the prediction. Other methods either fail or provide noisy features.}
   \label{fig:hard}
\end{figure*}

 % 
Saliency methods aim at helping the user to understand why a DNN made a particular decision by explaining the entire model. However, we observe two considerable limitations of these methods. First, they cannot explain the inner function of the neural system being examined. That means how internal neurons interact with each other to reach a particular prediction. As reported in \cite{200407213B}, it is difficult to verify claims about black-box models without explanations of their inner workings. 
A second limitation, they are susceptible to noise and spurious correlations. Whether due to a property of the DNN system obtained during the training phase (e.g., biased inputs, overfitting, or misspecification) or the method being used to capture saliency \cite{noisysal19, leavitetal20} as shown in Fig. \ref{fig:hard}). Alternatively, some methods seek to visualize the behavior of specific neurons \cite{olah_feature_2017} but cannot provide clear insights due to their large number and overall complex architectures. 
\par
In this paper, we propose a novel method that addresses the above limitations through the angle of causality. We show that a technique grounded in the theory of causal inference provides robust and faithful interpretations of model behavior while being able to reveal its neural interactions. Inspired by neuroscience, we analyze individual neurons' effects on model prediction by intervening in their connections (model's weights or filters).
%
%However, traditional explanation methods focus on correlation-based tools which lead to generate unstable and noisy explanations . This limits the ability of these methods in interpreting the system's behaviour when it gives wrong prediction for hard inputs (see example in Fig. \ref{fig:hard}). This, in turn, makes it difficult for users to rely on such methods and adopt them for critical decisions. Recently, causal analysis has emerged as a promising tool for DNN interpretability \cite{vig_causal_2020, geiger_causal_2021, cxplain_2019}. Here, we use causal analysis to address the limitations of standard attribution methods and propose a novel method to capture robust and faithful explanations. The motivation of our work is not limited to provide stable explanations, but to provide a method that allows understanding the inner working of trained DNNs by focusing on neural connections inspired by neuroscience.
\par
We summarize our contributions as follows. a) We propose a robust interpretability approach to capture meaningful semantics and explain the inner working of DNNs. b) Our methodology relies on path interventions and cause-effect relations, providing stable and consistent explanations. More specifically, we seek to answer questions such as: \textbf{\textit{would the model's prediction have been higher if we prevented the flow of signals through particular paths?}} or, \textbf{\textit{what would have been the decision of the model had we attenuated or removed an individual or a set of components at a particular layer?}}. Our analysis will lead to locating and isolating relevant and necessary information strongly and causally connected to model prediction up to a test of significance. c) We apply our method to vision models trained to classify MNIST, CIFAR10, and ImageNet data. d) We provide a flexible framework that can be applied to complex architectures and other tasks beyond interpretations. 
%
%Using test samples, our method estimates causal explanatory graphs for each class of interest and captures human-level semantics either from individual or aggregated responses to causal filters.
%////////////////////////////////////////////////////////////////////////////////////////////////////////////////////////////////////////////////////////////////////////////////////////////////////////////////////////////
%////////////////////////////////////////////////////////////////////////////////////////////////////////////////////////////////////////////////////////////////////////////////////////////////////////////////////////////
%%%%%%%%%%%%%%%%%%%%%%%%%%%%%%%%%%%%%%%%%%%%%%%%%%%%%%%%%%%%%%%%%%%%%%
% I NEED TO CHECK THE LAST SENTENCE AND WHICH ONE TO USE????????
\section{Related Work}
\label{rel_w}
\paragraph{Interpretability and Attribution Methods.} 
Interpretability for deep neural networks aims to provide insights into black box models' behavior. A broad family of methods has been developed in the past few years. The most common techniques are attribution methods which assign scores to input features indicating the contribution of each one to the model prediction. Gradient-based methods \cite{IG, gradxinp, gradshape, deconnet, Guidedback_2015, deeplift_2017} propagate gradients of pre-trained models from output backward until input. Recent studies have pointed out that these methods produce noisy and unstable attributions \cite{noisysal19, robustness}. Perturbation-based methods \cite{rise, occlusion, lime} are alternatives that focus on correlations between local perturbations of raw inputs and model output. They are black-box methods in the sense that they don't require access to the inner state of the model. Beyond these widely used techniques, various interpretability methods have been proposed \cite{surveyxai21}. Close to our work is \cite{CNNKnoledgeGraph}. They suggest disentangling knowledge hidden in the internal structure of DNNs by learning a graphical model. Their work focuses on convolutional neural networks (CNNs), where they fit the activations between neighboring layers. Our approach differs in what it considers explanatory graphs and how it infers them. We rely on causal analyses, which have been recently considered as an effective tool for DNN interpretability and explainability. Our framework does not assume a specific type of neural network, which makes the approach generic and flexible.
\paragraph{Capturing Explanations with Causality.}
More recently, causal approaches have been considered for interpreting DNNs. The inner structure of DNN has been viewed, for the first time, as a structural causal model (SCM) in \cite{chattopadhyay19a}. They use SCM to develop an attribution method that computes the causal effect of each input feature on the output of a recurrent neural network. Other causal approaches were specifically developed to explain NLP-based language models, such as causal mediation analyses \cite{vig_causal_2020} and causal abstraction \cite{geiger_causal_2021}. In contrast, \cite{cxplain_2019} have developed a model-agnostic approach (CXPlain) to estimate feature importance for model interpretations. They use a causal objective to train a separate supervised model (U-net) to learn causal explanations for another black-box model. 
%In addition to being compatible with any black-box model, this method provides uncertainty estimates of explanations via bootstrapping. 
An important limitation of this method is that it has to be trained to learn to explain the target model. Another point is that its causal property is limited to the extrinsic effect of input on causing a marginal change in output. Therefore, it cannot link explanations to the model’s internal structure, which remains a black box. 
%////////////////////////////////////////////////////////////////////////////////////////////////////////////////////////////////////////////////////////////////////////////////////////////////////////////////////////////
%////////////////////////////////////////////////////////////////////////////////////////////////////////////////////////////////////////////////////////////////////////////////////////////////////////////////////////////
%%%%%%%%%%%%%%%%%%%%%%%%%%%%%%%%%%%%%%%%%%%%%%%%%%%%%%%%%%%%%%%%%%%%%%
\section{Causal Graph Inference of Neural Networks}
\label{sec:ibbe}
%\subsection{Intuition from Signal Processing}
\subsection{Notation and Intuition} 
\label{sec:not}
\begin{figure*}[t]
   \centering
   \includegraphics[width=0.9\linewidth]{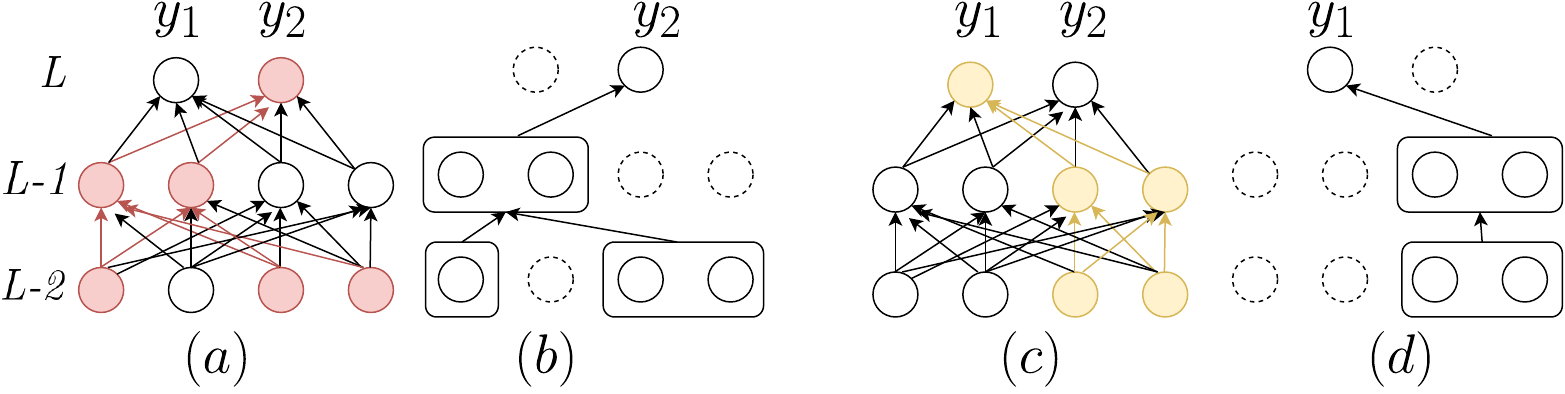}
   \caption{\textbf{Causal connections within the last three layers of a neural network.} (a) and (c) Coloured paths (red/yellow) transpose signals between layers to labels $y_{1}/y_{2}$, respectively. (b) and (d) Two abstract graphs are obtained by causal inference. In each graph, neutral neurons (marked by dots) hold variant information which don't influence models' behavior for the corresponding label.}
\label{fig:1}
\end{figure*}

\paragraph{Notation} We denote by $\boldsymbol{x}$ an input image (without loss of generality), and $y \in \mathbb{R}^{n_{y}}$ its corresponding output label, where $n_{y}$ is the number of classes. We also denote by $\hat{y}\in \mathbb{R}^{n_{y}}$ its predicted output obtained by a pre-trained neural network $N(L)$ composed of $L$ layers. We define the relation ${l \rightarrow l+1}$ to refer to directed edges or connections between hidden nodes of layers $l$ and $l+1$, respectively. At every hidden node $j$ of the $l$-th layer, we define features or activation map $a^{l}_{j}$. We denote by causal graph $\mathcal{G}$ an abstraction of $N(L)$ as shown in Fig.\ref{fig:1} (b) and (d). We use the term explanatory graph to refer to causal graphs whose nodes hold important features.
\paragraph{Intuition} The activated signals $\boldsymbol{a}^{l}$ flow to the next layer $l+1$ through weighted edges $\boldsymbol{W}^{l \rightarrow l+1}$ connecting hidden nodes of layers $l$ and $l+1$. These weights control the strength of information flow between two layers in a manner physically analog to a switch. In physical systems, manipulating the state of a switch (e.g., on-off or via continuous interventions) would change the system's physical state, thereby providing an interpretation of its behavior.  
We set this intuition to motivate our work. To our knowledge, there is relatively little research on DNN explainability by manipulating weights.  
%
%which bases on a simple idea yet, to our knowledge, has taken little attention in the literature of deep learning explainability.
%
\subsection{Problem Formulation}
\label{sec:prob}
Our goal is to discover causal explanatory graphs of $N(L)$ via path (equiv. weights) interventions. %
\begin{figure*}[tb]
    \centering
    \includegraphics[width=\textwidth]{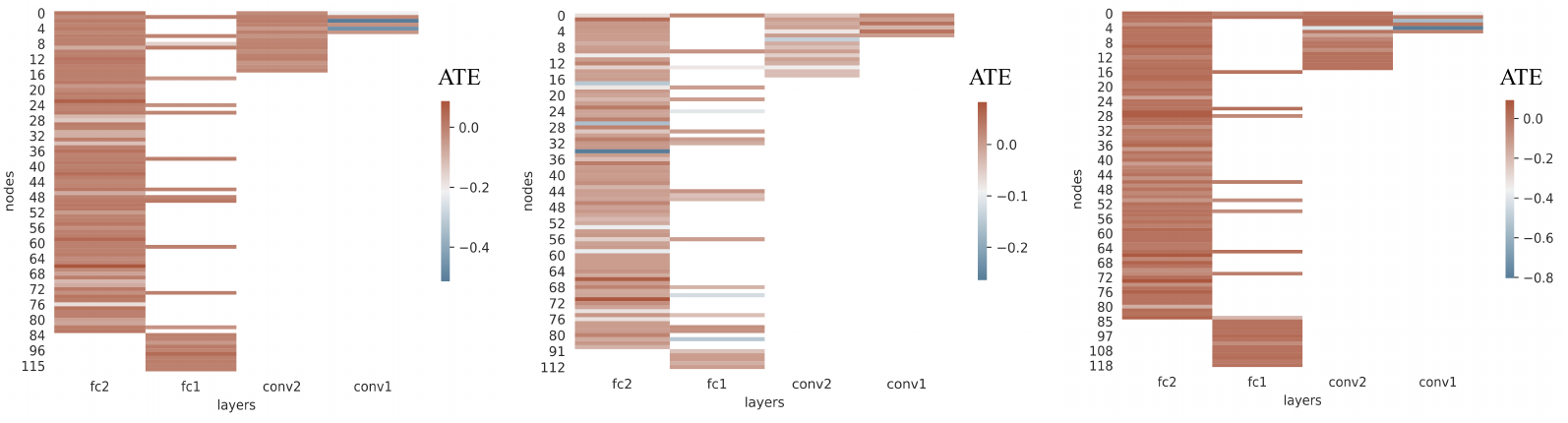}
    \caption{\textbf{Heatmap of the Average Treatment Effect (ATE).} We show the effect of path interventions on the convolution and linear layers of LeNet architecture for digits $3$, $7$ and $8$ respectively. Y-axis indicates the total number of nodes overall layers. For instance, conv1 has 6 nodes (channels) and the last hidden layer $fc_{2}$ has 84 nodes. The colorbar indicates the relative ATE values w.r.t the original outputs.}
    \label{fig:te}
\end{figure*}
% IN progress
%When the switch at ON state, it means that the path is open and the signal transmits from $i$ to $j$, and when it becomes OFF, the path is closed and $i$, $j$ are disconnected. 
%
Formally, we set the problem as follows. Let $\boldsymbol{W}^{l \rightarrow l+1}\in\mathbb{R}^{n_{c}\times n_{p}}$ be the weight matrix of the directed edges from layer $l$ to $l+1$, where $n_{p}$ is the number of parent nodes in $l$ and $n_{c}$ is the number of child nodes in $l+1$. These nodes define a sub-graph $\mathcal{G}^{l \rightarrow l+1}$. 
%For simplicity, we will omit the superscript ${l \rightarrow l+1}$ when we focus on a pre-defined parent layer $(l)$. 
Let $\boldsymbol{w}^{l \rightarrow l+1}\in \mathbb{R}^{n_{t}\times n_{p}}$, be the paths connecting $n_{p}$ nodes in $l$ to $n_{t}$ target nodes in $l+1$ ($n_{t} < n_{c}$). Our problem is then to estimate how significant the causal or treatment effect $(TE)$ resulting from intervening on the weights $w^{l \rightarrow l+1}_{j}$ at node $j$: 
\begin{equation}
\label{eq:1}
\boldsymbol{P}\{TE(do(w^{l\rightarrow l+1}_{j}); \boldsymbol{\hat{Y}}, \boldsymbol{X}, \boldsymbol{W}\setminus w^{l\rightarrow l+1}_{j}) = 0 \} < \alpha,
\end{equation}
where $do(w^{l\rightarrow l+1}_{j})$ is a mathematical operation referring to the action of interventions. $\boldsymbol{X}$ is a subset of inputs in the data manifold $\mathcal{\boldsymbol{X}}$, and $\boldsymbol{\hat{Y}}$ are their predictions (pre-softmax layer). $\alpha>0$ is a probability threshold (equiv. p-value) that measures "significance". The formula in (\ref{eq:1}) defines a form of hypothesis testing, where the null hypothesis states that interventions on the paths from node $j$ will not affect or change the original predictions of the model $\boldsymbol{\hat{Y}}$. This formula means rejecting the null hypothesis, and will lead us to identify the most influential nodes of $l$ on $\boldsymbol{\hat{Y}}$.  
%////////////////////////////////////////////////////////////////////////////////////////////////////////////////////////////////////////////////////////////////////////////////////////////////////////////////////////////
%+++++++++++++++++++++++++++++++++++++++++++++++++++++++++++++++++++++++++++++++++++++++++++++++++++++++++
\subsection{Causal Inference} 
\label{sec:ci}
In this section, we provide the details of our methodology for solving (\ref{eq:1}). We focus on vision models which encompass a set of convolution and MLP layers. Specifically, we use LeNet \cite{lenel} with MNIST data for ease of explanations. The experiments section shows applications on common datasets and more complex architectures. Here, we seek to capture the causal explanatory graph of LeNet given inputs of digit $k\in \mathbb{N}_{9}$.    
\paragraph{Treatment Effects} The first step of our approach is to compute the effects of path interventions on model outputs.
%caused by interventions on paths directed from node $j$ in hidden layer $l$. 
Let us consider the MLP example in Fig. \ref{fig:1} (a) and (c). 
The interventions on the paths in the last hidden layer $L-1$ allow measuring the effect on the outputs ($y_{1}$ and $y_{2}$) directly. Meanwhile, for layer $L-2$, the effects of interventions are mediated by the responses of hidden neurons in descendant layers; in this example, the child layer $L-1$. The strength of response to path interventions depends on the structure and complexity of neural networks. Our goal is thus to analyze how significant these effects are. First, we define the treatment effect as a measure of the  difference corresponding to path interventions.
\begin{definition} (\textbf{Treatment Effect}) Let $\boldsymbol{X}$ be a set of input features and $\hat{\boldsymbol{Y}}$ the corresponding output of a neural network $N(L)$. Let $w^{l \rightarrow l+1}_{j}\in \mathbb{R}^{n_{t}}$ be the weights vector directed from node $j$ in layer $l$ to $n_{t}$ nodes in layer $l+1$. By holding all other weights $\boldsymbol{W}\setminus (w^{l \rightarrow l+1}_{j})$ fixed and intervening on $w^{l \rightarrow l+1}_{j}$ (i.e., $do(w^{l \rightarrow l+1}_{j})$), we define their effect as follows:
%\begin{equation}
%\begin{split}
%&TE(do(w^{l \rightarrow l+1}_{j}); \hat{\boldsymbol{Y}}, \boldsymbol{X}, \boldsymbol{W}\setminus w^{l \rightarrow l+1}_{j}) = \\
%&\frac{\hat{\boldsymbol{Y}}_{w^{l \rightarrow l+1}_{j}=u_{1}}(\boldsymbol{X} ) - \hat{\boldsymbol{Y}}_{ w^{l \rightarrow l+1}_{j}=u_{0} }(\boldsymbol{X}) }{ \hat{\boldsymbol{Y}}_{w^{l \rightarrow l+1}_{j}=u_{0}} (\boldsymbol{X} ) }
%\label{eq:tce}
%\end{split}  
%\end{equation}
%\end{definition} 
%
\begin{equation}
\begin{split}
&TE(do(w^{l \rightarrow l+1}_{j}); \hat{\boldsymbol{Y}}, \boldsymbol{X}, \boldsymbol{W}\setminus w^{l \rightarrow l+1}_{j}) = \\
&\hat{\boldsymbol{Y}}_{w^{l \rightarrow l+1}_{j}=u_{1}}(\boldsymbol{X} ) - \hat{\boldsymbol{Y}}_{ w^{l \rightarrow l+1}_{j}=u_{0} }(\boldsymbol{X})
\label{eq:tce}
\end{split}  
\end{equation}
\end{definition} 
where $u_{1}$ and $u_{0}$ are intervention variables defined below.
%defined in Section \ref{sec:int}.  
Equation (\ref{eq:tce}) measures the relative change of the outputs distribution over the inputs $\boldsymbol{X}$ given the same actions at node $j$. By considering all the $n_{p}$ nodes in layer $l$, we obtain the set of distributions $\{ TE \}_{j=1,...,n_{p}}$. IFig. \ref{fig:te}, shows samples of the average treatment effect obtained over $\boldsymbol{X}$ when interventions correspond to removing edges in the hidden layers. 
\paragraph{Test of Significance} To capture the most influential nodes in the parent layer $l$, we consider hypothesis testing as formulated in eq. (\ref{eq:1}). We observe that the null distribution is approximately Gaussian, given the sufficiently large number of samples (in training sets). This makes the z-test an appropriate choice to solve the problem. We set the probability threshold $\alpha$ to its common value $0.05$. That means, the effect of intervening on the paths coming out from node $j$ is significant when eq. (\ref{eq:1}) holds with $5\%$ chance of error.
%Since we have $n_{p}$ distributions, each corresponding to a hypothesis, we use multiple comparisons.
%More specifically, we use Holm-Bonferroni method \cite{holme} to adjust the rejection criterion for each hypothesis given a significance level $\alpha$. We use the common value $\alpha=5\%$. 
%Then, the effect of intervening on the paths coming out from node $j$ is significant when Eq. (\ref{eq:1}) holds given 0.05 chance of error. 

%Note that Eq. (\ref{eq:1}) allows finding a threshold for positively significant effects. We observe that the distribution under null hypothesis is symmetric and normal, so we use a signed threshold to detect the influential nodes for significantly positive and negative effects.
%
\paragraph{Path Interventions} Following the intuition of our work, we propose the interventions $w^{l \rightarrow l+1}_{j}=u$ such that 
$u=\beta w^{l \rightarrow l+1}_{j}$, where $\beta$ can either be discrete (remove connections) or continuous (attenuate connection's effect). In the discrete case, $\beta$ is binary so that $u_{1} = 0$ and $u_{0} = w^{l \rightarrow l+1}_{j}$. In the continuous case, we propose to sample $\beta$ from a uniform distribution $\mathcal{U}(b-\epsilon,b+\epsilon)$, where $\epsilon< b < 1.0$ is a predefined parameter and $\epsilon=0.01$. We use continuous interventions to evaluate the consistency of the causal effects and estimated graphs. 
%////////////////////////////////////////////////////////////////////////////////////////////////////////////////////////////////////////////////////////////////////////////////////////////////////////////////////////////
%+++++++++++++++++++++++++++++++++++++++++++++++++++++++++++++++++++++++++++++++++++++++++++++++++++++++++
\subsection{Path Selection} 
\label{sec:sel}
\begin{figure}[!h]
\centering
\includegraphics[width=0.4\textwidth]{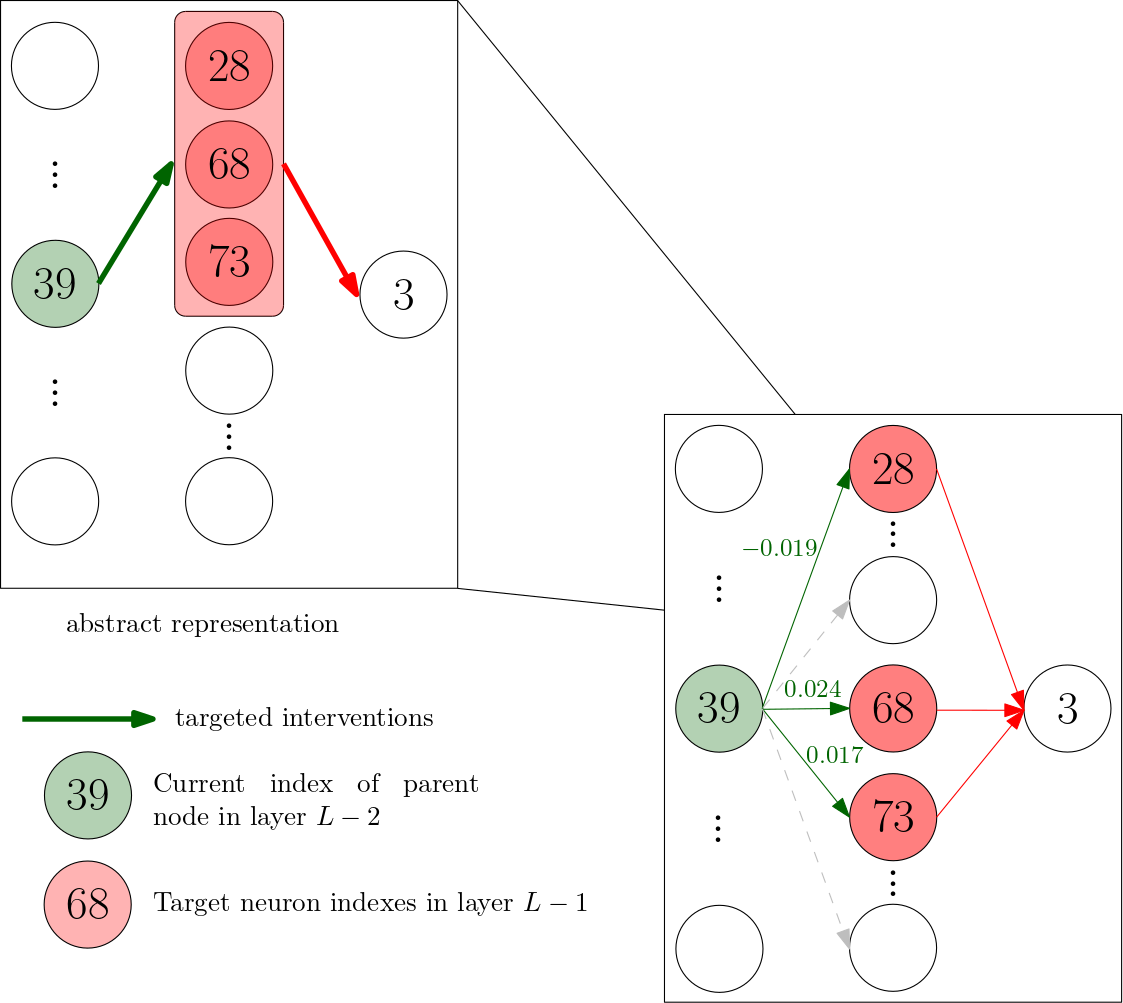}
\caption{Path selection.}
 \label{fig:lenet-3}
\end{figure}

So far, we explained how to solve (\ref{eq:1}) using $w^{l\rightarrow l+1}_{j}$ for each parent node $j$. These weights correspond to a subset of targets we identify here via path selection. Indeed, manipulating all possible connections for a node $j$ given parent layer $l$ is computationally expensive and intractable for complex architectures with many neurons. An efficient way is to pay attention to specific paths and nodes via selection criterion. We propose a top-down approach starting from a specific output (e.g., class). It implies  sequential processing starting from the last layer until reaching layer $l$. Let us consider we seek to compute the effects of path interventions of LeNet's layer $L-2$ for digit $3$ (as shown in Fig. \ref{fig:lenet-3}). 
%
%\begin{wrapfigure}[17]{r}{0.5\textwidth}
% -- OLD
%We start with the paths directed from all nodes in the parent layer $(L-1)$ to node $3$ of the output layer $L$. Computing eq. (\ref{eq:1}) gives the causally significant nodes in $(L-1)$ (the red nodes in this example). 
%Nodes whose removed edges decrease significantly model's performance are considered necessary for that output (i.e., $TE < -\tau$ up to $\alpha$). Nodes whose removed edges improve predictions are considered noisy or distracting ($TE > \tau $ up to $\alpha$). 
% NEW
We start with the paths directed from all nodes in the parent layer $(L-1)$ to node $3$ of the output layer $L$. Computing eq. (\ref{eq:1}) reveals the most relevant nodes in $(L-1)$, up to a significance test $\alpha=5\%$. To identify the impact of these nodes on the model, we must look at the behavior of causal effects. Negative values explain a drop in class prediction when removing edges or amortizing weights, while positive values explain an improving prediction. Hence, the nodes revealed when the causal effect is significantly below zero are considered necessary for that output (the red ones in this example). In contrast, we discover noisy or distracting nodes when path interventions have a significant positive effect.
We thereby select the necessary (red) nodes as targets for the next sub-graph $\mathcal{G}^{L-2 \rightarrow L-1}$. We repeat the same process on $L-2$, but this time we simultaneously intervene on all paths directed from a parent node (green node) to the targets. With this process, we can efficiently estimate relevant nodes in all intermediate layers while focusing on meaningful interventions. Algorithm \ref{alg:1} shows the implementation steps for discovering the causal explanatory graphs of a classification neural network. We provide some visualizations of LeNet's causal graphs in the supplementary. 
%
%
%\begin{wrapfigure}[20]{O}{0.65\textwidth}
%\vspace{-2em}
%\begin{figure}[!h]
%\begin{minipage}{.9\linewidth}
\begin{algorithm}[tb]
\caption{Causal Explanatory Graph Inference ($\mathcal{G}$) of a DNN} 
\label{alg:1}
\textbf{Input}: $N(L)$ pre-trained DNN, $\boldsymbol{W}$ weights, $\boldsymbol{X}$ task-specific examples, $\hat{\boldsymbol{Y}}$ model outputs, $(k)$ task index \\
\textbf{Output}: $\mathcal{G}$ (Dict. of important nodes and their relations), $\mathcal{D}$ (Dict. of irrelevant nodes) 
\begin{algorithmic}
\STATE $l \leftarrow L-1$, $\beta \leftarrow \{0, 1\}$
\WHILE{$l>0$}
\STATE $n_{p} \leftarrow dim(l)$
\FOR{$j=1$ to $n_{p}$}
\STATE $u \leftarrow \beta w^{l \rightarrow l+1}_{j}$
\STATE $do(w^{l \rightarrow l+1}_{j}=u)$
\STATE Compute $TE(do(w^{l \rightarrow l+1}_{j}), \hat{\boldsymbol{Y}}, \boldsymbol{X}, \boldsymbol{W})$ for all  $\boldsymbol{X}$
\STATE Solve (\ref{eq:1}) and get nodes $(J^{l}, I^{l})$
\STATE $\mathcal{G}^{l \rightarrow l+1} \leftarrow J^{l}$, $\mathcal{D}^{l \rightarrow l+1} \leftarrow I^{l}$
\ENDFOR
\STATE $l \leftarrow l - 1$
\ENDWHILE
\end{algorithmic}
\end{algorithm}
%////////////////////////////////////////////////////////////////////////////////////////////////////////////////////////////////////////////////////////////////////////////////////////////////////////////////////////////
%////////////////////////////////////////////////////////////////////////////////////////////////////////////////////////////////////////////////////////////////////////////////////////////////////////////////////////////
%%%%%%%%%%%%%%%%%%%%%%%%%%%%%%%%%%%%%%%%%%%%%%%%%%%%%%%%%%%%%%%%%%%%%%
\section{Explanations from Causal Graphs}
\label{sec:feat}
\begin{figure}
    \centering
    \includegraphics[width=0.5\textwidth]{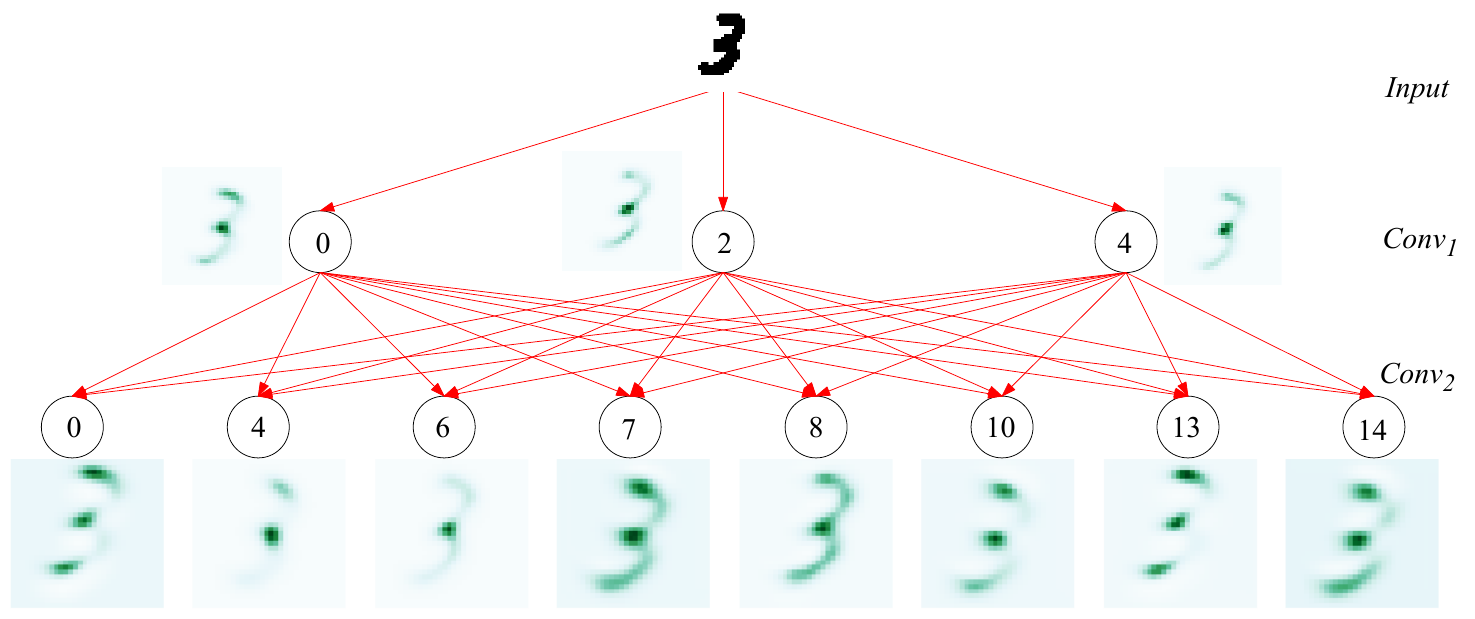}
    \caption{\textbf{Illustration of LeNet's causal sub-graphs $\mathcal{G}^{input\rightarrow conv_{1}}, \mathcal{G}^{conv_{1}\rightarrow conv_{2}}$ for class $3$.} The resulted attributes provide visual interpretations for a sample image correctly classified by the model. They are up-sampled and normalized to reflect pixel-wise probabilities (Dark greens correspond to peaks with highest scores.).}
    \label{fig:attr3}
\end{figure}

%Attributions and object semantics can be extracted in a robust way by taking advantage from the hierarchical structure of the estimated causal graphs of DNN.
The hierarchical structure of the causal graphs enables robust extraction of attributions and high-level semantics. Instead of capturing a single saliency map from all activations, we rely on features response along the causal pathways. We empirically show that these features are more stable and consistent compared to traditional attribution methods. As reported in \cite{noisysal19}, the reason for these methods to produce noisy and unstable attributions is due to distracting features in DNNs. Our method can remove the features that negatively affect model's prediction, and isolate important neurons in causal graphs/sub-graphs. Formally, given the sub-graph $\mathcal{G}^{l \rightarrow l+1}$, we extract salient interpretations ($s^{l+1}_{i}$) at a node $i$ in $l+1$ as follows  
\begin{equation}  
s^{l+1}_{i} = \frac{1}{J^{l}}\sum^{J^{l}}_{j=1} f( w^{l \rightarrow l+1}_{ji}, a^{l}_{j}), 
\label{eq:sal}
\end{equation}
where $a^{l}_{j}$ is the j-th activated signal of layer $l$, $J^{l}$ is the number of parent nodes in $l$ connected to the child node $i$ in the layer $l+1$. The response $f$ depends on the structure of the parent layer. For convolution layers, $w_{ji}$ is a filter and $f$ is a convolution function; whereas for MLPs, $f$ is linear function. Fig. \ref{fig:attr3} shows causal sub-graphs, up to \emph{conv2} layer (for visualization), and the underlying attributions for a LeNet model successfully classified its input. 
%Further results on ImageNet are in supplementary materials.
\par
\begin{figure}[t]
    \centering
    \includegraphics[width=0.45\textwidth]{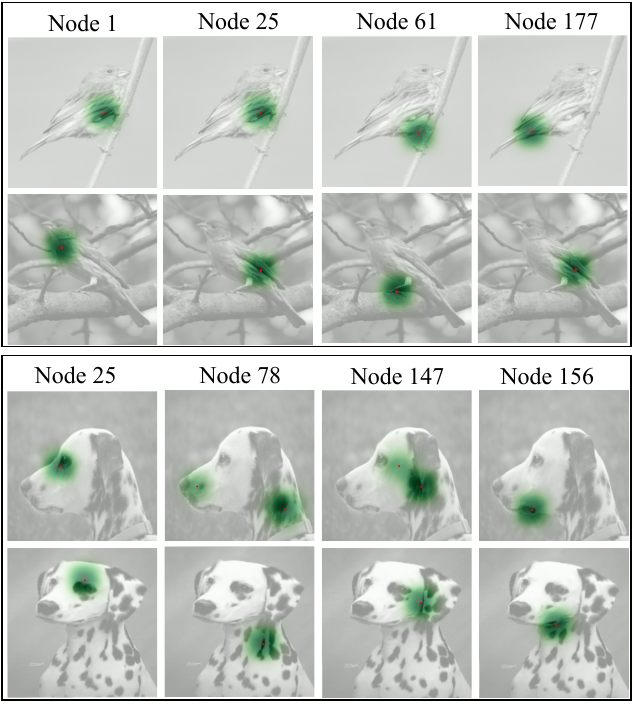}
    \caption{\textbf{Visualizing explanations obtained by the top-1 causal filters.} We show four examples for two object classes (from ImageNet). Important neurons belong to the causal sub-graph connecting the last Conv layers $l=layer4.1.conv1$ and $l+1 = layer4.1.conv2$ of ResNet18. We can observe consistent attributes for similar inputs. The red point indicates the location of the peaks corresponding to the absolute maximum response.}  
    \label{fig:filters}
\end{figure}

 
Note that eq. $(\ref{eq:sal})$ aggregates at every node $i$ the responses of its parent nodes to the filters/weights. We may also be interested in analyzing and interpreting the role of each filter between the pairs ($i, j$). Fig. \ref{fig:filters} is an example of the response  to the top-1 filters (w.r.t. the amplitude of their causal effects) for a set of relevant nodes in the last convolution layer of ResNet18. 
Causal attributes (of object parts) are refined by extracting the response's local maxima (and minima). 
%////////////////////////////////////////////////////////////////////////////////////////////////////////////////////////////////////////////////////////////////////////////////////////////////////////////////////////////
%////////////////////////////////////////////////////////////////////////////////---EXPERIMENTS---////////////////////////////////////////////////////////////////////////////////////////////////////////////
%%%%%%%%%%%%%%%%%%%%%%%%%%%%%%%%%%%%%%%%%%%%%%%%%%%%%%%%%%%%%%%%%%%%%%
\begin{figure*}
    \centering
    \includegraphics[width=\textwidth]{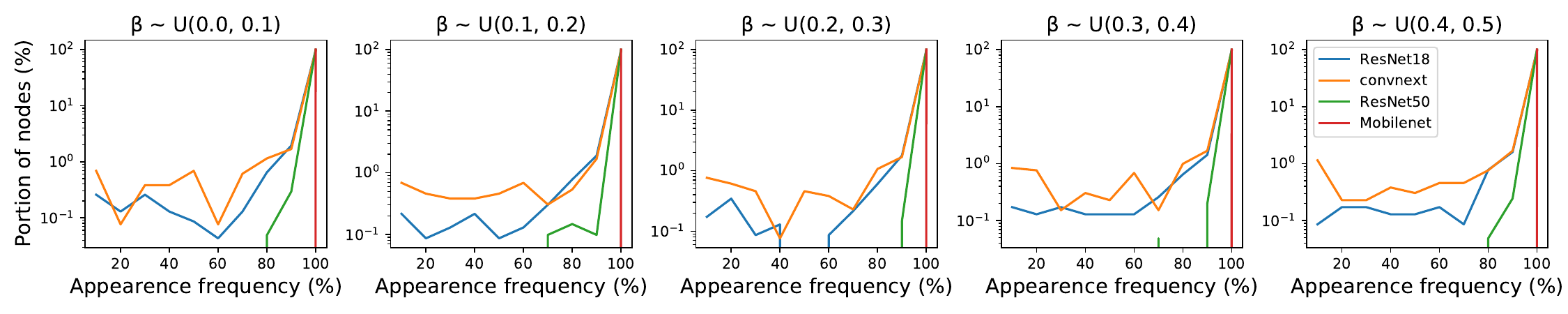}
    \caption{\textbf{Reliability assessment of causal graphs.} We show results on four complex architectures. On the x-axis, we show the frequency of appearance of a node (\%). On the y-axis is the portion of all the nodes appearing at least once during the experiments (\%).}
    \label{fig:table_stability}
\end{figure*}
\section{Experiments}
\label{sec:exp}
The experiments section splits into two parts: 1) we evaluate our algorithm's capacity to estimate stable and consistent causal graphs; 2) we evaluate the explanations captured by causal graphs and compare them to various attribution methods using standard explanation metrics. 
\paragraph{Models and datasets}
We evaluate our method on the LeNet model trained on MNIST data and the following architectures: ResNet18 \cite{7780459}, ResNet50V2 \cite{resnetv2},  MobileNetV2 \cite{8578572}, and on the latest architecture ConvNext \cite{connext}; the tiny version. 
These models were trained on the large-scale ImageNet data (ILSVRC-2012) \cite{ILSVRC15}. We also fine-tuned these architectures on CIFAR10 dataset after updating their last classification layer. We divide the validation sets into validation and test sets. We use the samples in validation sets to discover causal explanatory graphs and the test set for evaluating the explanations.
%
%To conduct our experiments, we choose the following datasets: CIFAR10, and miniImageNet split into evenly distributed train and test images (with original resolution ($224\times 224$)) across 100 classes. For miniIMageNet, we rearranged the samples to ensure they are in-distribution. 
%
%
%We updated the last classification layer and fine-tuned the models w.r.t. the underlying datasets. Finally, we used the samples in test sets to generate class-specific causal explanatory graphs. Further details on data and implementations are reported in Appendix.
%
%
\paragraph{Comparison methods} We selected the most popular attribution methods from two categories: model-agnostic (black-box) and gradient-based (white-box) methods. We chose RISE \cite{rise} and Occlusion \cite{occlusion} as black-box methods, and the following gradient-based methods: Integrated-Gradient (IG) \cite{IG}, Saliency \cite{saliency}, Gradient Shape \cite{gradshape}, GradXInput \cite{gradxinp}, DeconvNet \cite{deconnet} and Excitation Backprob (MWP) \cite{exbackprob}. %Also, we provide sanity check experiments with common baselines in Appendix.
%For the NLP experiments, we ...
\subsection{Evaluating the Reliability of Causal Graphs}
% For the causal graphs, we introduce robustness metric to evaluate how consistent are the graphs when varying the intervention parameter $\beta$.
%  
In this experiment, we evaluate the stability and consistency of our estimation of causal graphs. 
Since the causal effect is based on path interventions, we need to ensure consistency in the statistical test results no matter what intervention values are used (i.e., binary or continuous). We do so by running $1000$ experiments with an intervention parameter randomly sampled from a uniform distribution $\mathcal{U}(b-0.01,b+0.01)$. Here, $b$ changes monotonically every $10$ runs in the range $(0.01, 0.5)$. We report reliability by measuring the frequency of detecting the same important nodes in each layer (in percentage). In Fig \ref{fig:table_stability}, we show for a few samples of $\beta$ the distribution of the nodes versus their appearance rate.  
As we can see, the stability of the graph does not rely on the value chosen for the intervention parameter. Regardless of the value of $\beta$, a considerable proportion ($98\%$ to $100\%$) of the nodes appear in every experiment.
The stability of causal graphs indicates two facts: (1) the importance of the activated signals, which are affected by weights attenuation. (2) our method is not sensitive to the choice of interventions (binary or continuous). Furthermore, the causal effect is significant even when reducing the strength of the signal along the causal path by only a factor of $1/2$. These results ensure that the properties of single neurons might indeed be representative of model's behavior.
% old methods could not ensure this because of the method itself
%
\subsection{Evaluation of causal explanations}
The causal graphs estimated by our method summarize knowledge from all hidden layers in the DNN and enable better interpretability. 
%Indeed, there are more than one node that can be used to interpret a DNN model. 
For example, Fig.\ref{fig:attr3} shows that for classifying digit $3$, there exist $8$ relevant nodes in the Conv2 layer, each encoding signal activated at different parts of the object. To compare the explanations obtained by our method with existing attribution methods, we aggregate attributions at the relevant nodes in a specific layer. 
%For end-users, sometimes one saliency corresponding to the entire model can be insufficient to explain its prediction on hard examples as the case shown in Fig. \ref{fig:hard}. More cases are shown in supplementary materials. 
Then, we evaluate the stability and faithfulness of explanations using standard state-of-the-art metrics. The evaluations are performed using the Quantus library \cite{hedstrom_quantus_2022}. Details on explanation metrics and attributions visualization are provided in the supplementary.
\begin{figure}
\footnotesize
\setlength{\tabcolsep}{1pt}
\begin{tabular}{cc}
\includegraphics[width = 0.5\linewidth]{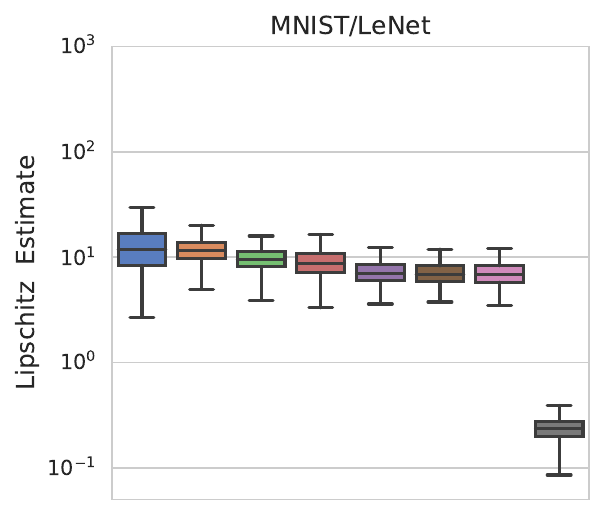} &
\includegraphics[width = 0.5\linewidth]{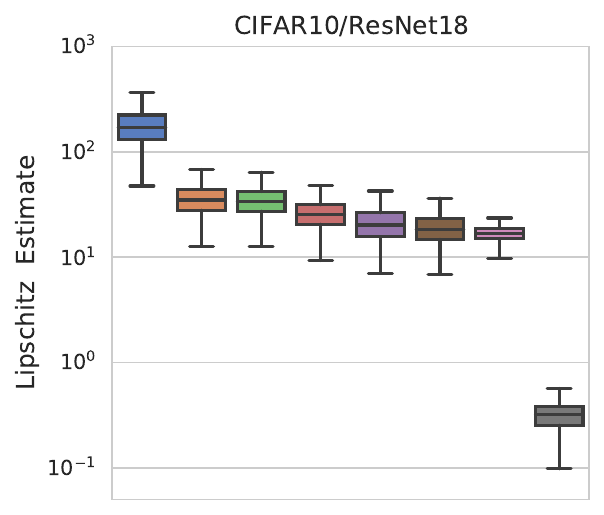} \vspace{-9pt}\\
%\hspace{-20pt}
\vspace{-12pt}
\includegraphics[width = 0.5\linewidth]{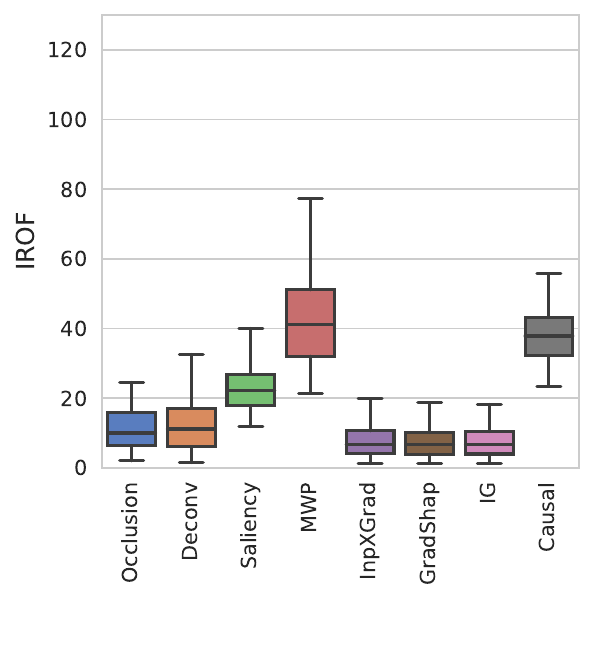} &
\includegraphics[width = 0.5\linewidth]{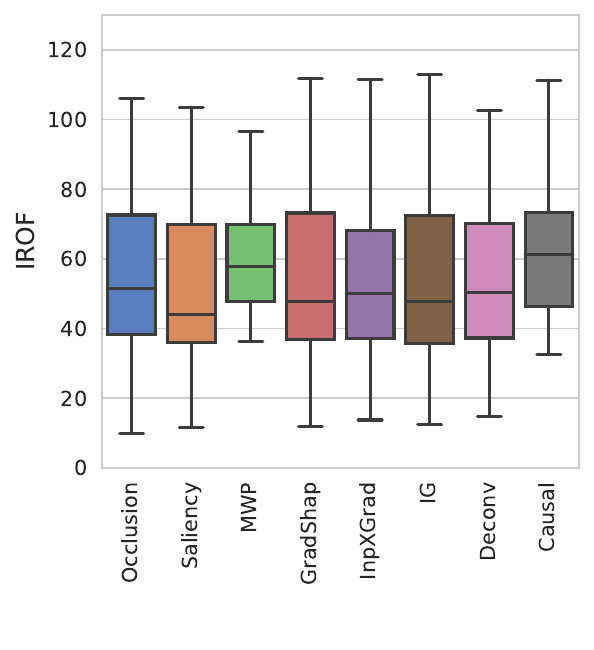}
\end{tabular}
\caption{\textbf{Quantitative evaluations of attribution methods for LeNet on MNIST and ResNet18 on CIFAR10.} For each metric, we compare 7 attribution methods to the causal explanations obtained by our method using test images. The bars show mean and variance over samples. Lower Lipschitz Estimates (w.r.t. means) indicate higher stability. Higher IROF values (w.r.t. means) indicate strong relation between explanations and predictions.}
\label{fig:metrics}
\end{figure}
\begin{figure}
\footnotesize
\setlength{\tabcolsep}{1pt}
\begin{tabular}{c}
\includegraphics[width = 1.1\linewidth]{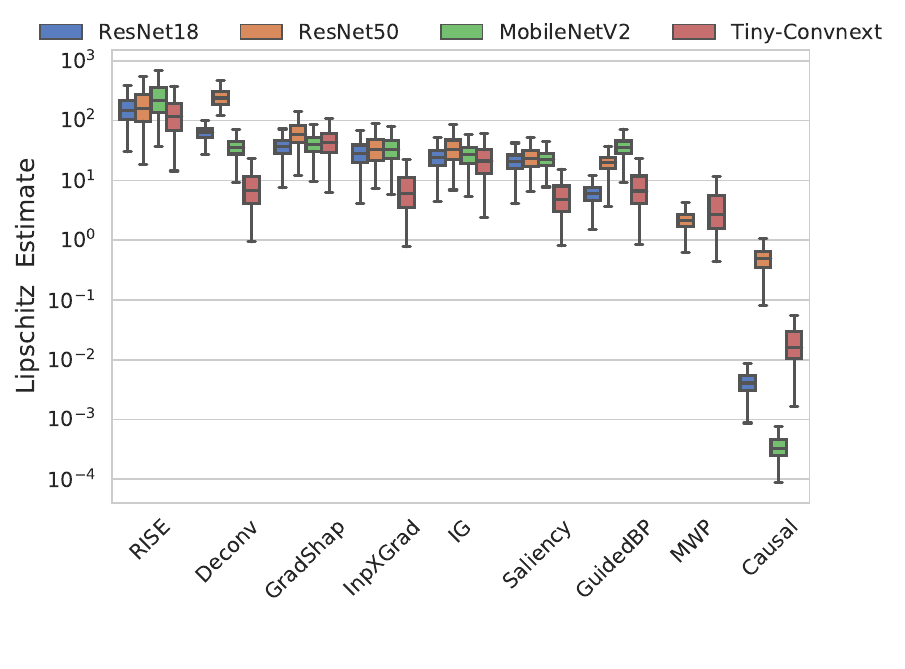} \vspace{-22pt}\\
\hspace{-20pt}
\vspace{-12pt}
\includegraphics[width = \linewidth]{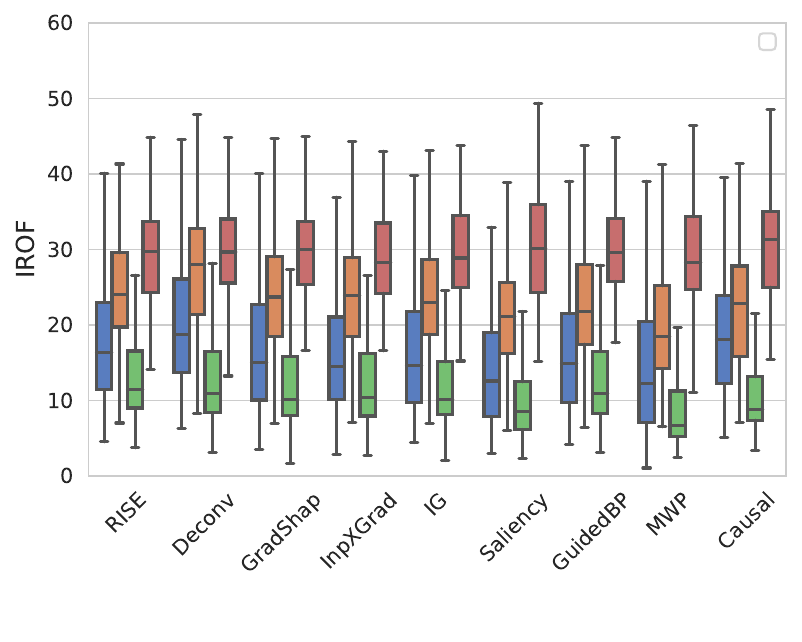}
\end{tabular}
\caption{\textbf{Quantitative evaluations of explanations for complex architectures on ImageNet.} We evaluate 9 methods including ours using 10 representative classes from the test set.}
\label{fig:imgnet_LE}
\end{figure}
%
%////////////////////////////////////////////////////////////////////////////////////////////////////////////////////////////////////////////////////////////////////////////////////////////////////////////////////////////
%+++++++++++++++++++++++++++++++++++++++++++++++++++++++++++++++++++++++++++++++++++++++++++++++++++++++++
\paragraph{Stability:}
Stability measures consistency of explanations against local perturbations of inputs. Here, we adopt Lipschitz Estimate (LE)~\cite{robustness}, which calculates the maximum variance between an input and its $\epsilon$-neighbourhood, where $\epsilon$ refers to the level of perturbations. We generate perturbations by adding white noise to inputs from the test sets. We compute explanations for every input in specific class and its noisy sample using the graphs estimated from the validation data. The maximum euclidean distance between explanations is then obtained over multiple runs where new perturbations are generated. Fig. \ref{fig:metrics} reports the results for LeNet trained on MNIST and ResNet18 fine-tuned on CIFAR10, and Fig.\ref{fig:imgnet_LE} shows the results for four different architectures trained on ImageNet data.
\par
The results (in Fig. \ref{fig:metrics} and \ref{fig:imgnet_LE}) clearly indicate that the explanations generated from the causal graphs are more stable and consistent compared to other attribution methods. The explanations generated by these methods show higher variance to perturbations depending on the dataset and model. In contrast, the explanation from causal graph show consistent stability. 
Our method has the lowest variance with significant margin compared to the best method in each experiment.
%Our method has the lowest variance with significant margin ($\Delta$) compared to the best method in each experiment: $\Delta = 6.57$ for LeNet on MNIST data, $\Delta = 7.83$ for ResNet18 on CIFAR10, and $\Delta = 0.72$, $\Delta=2.66$, for ResNet18 and ConvNext on ImageNet, respectively.     
% 
\paragraph{Faithfulness:} Evaluating attributions relevance for the decision obtained by the model is essential to ensure correctness and fidelity of explanations. This is commonly done by measuring the effect of obscuring or removing features from the input on model's prediction. Different techniques have been proposed to score the relevance of explanations \cite{bhatt_evaluating_2021, robustness, monoticity2019, Bach2015OnPE, irof}. Here, we used iterative removal of features (IROF) \cite{irof}. An image is partitioned into patches using superpixel segmentation. The patches are sorted by their mean importance w.r.t the attributions in each patch. At every iteration, an increasing number of patches with highest relevance are replaced by their mean value. The IROC computes the mean area above the curve for the class probabilities (perturbed vs. original predictions). We applied this metric to evaluate each explanation method including ours.
Fig. \ref{fig:metrics}) shows that our method outperforms other methods and is comparable to MWP \cite{exbackprob} (with a relatively small margin between their medians). For ResNet18 trained on CIFAR10, most attribution methods show higher scores than LeNet on MNIST. Furthermore, the explanations obtained by our method and MWP show less sensitivity to the different data and models, indicating better trustworthiness. Fig. \ref{fig:imgnet_LE} shows IROF results for different architectures trained on ImageNet. On ImageNet, all methods, including ours, agree on the differences in behavior between the four models and that ConvNeXt is more trustable than standard ConvNets. For interested readers, we refer to \cite{connext} for further details about the core design of the ConvNeXt family of architectures.
%
%For our case, we also measured the alignment between model performance and important neurons from causal graphs (Fig. \ref{fig:}) to show how these graphs accurately reflect the underlying behavior.
% I need to emphasize that this is not a pruning
%
%
\subsection{Fidelity of class-specific causal neurons}
\begin{figure}
\centering
\footnotesize
\setlength{\tabcolsep}{1pt}
\begin{tabular}{cc}
\includegraphics[width=0.5\linewidth]{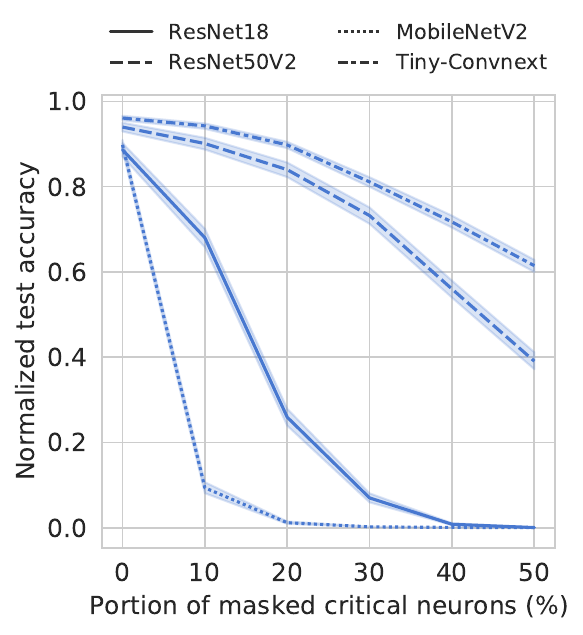} &
\includegraphics[width=0.5\linewidth]{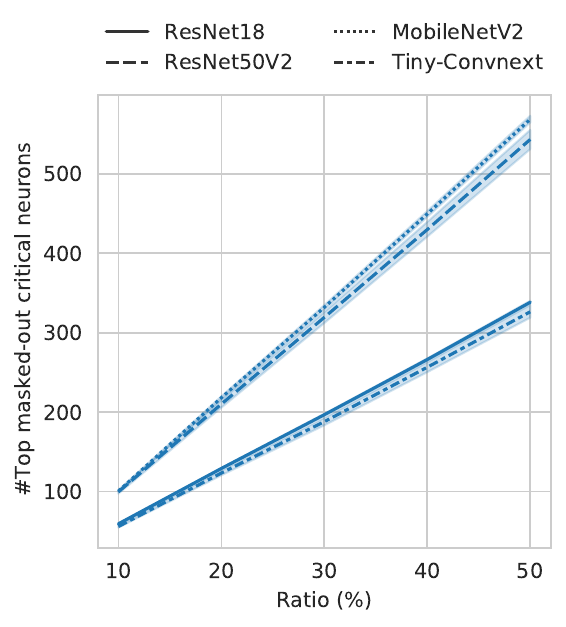} 
\end{tabular}
\caption{\textbf{Fidelity of class-specific causal neurons to the model.} The left figure shows the test accuracy of four models when masking out the top-k ($\%$) portion of causal neurons discovered as critical using our path interventions method. The figure shows the average accuracy over ten representative classes selected from ImageNet. The right figure shows the absolute number of critical neurons at each portion.}
\label{fig:fid}
\end{figure}

The causal neurons discovered as critical (or relevant) through interventions should accurately describe model behavior. We evaluate this by measuring the model accuracy on a specific class when masking out the critical neurons connected to this class. That means high-fidelity neurons should cause a drastic drop in accuracy under discarding them. We illustrate this behavior on four models trained on ImageNet in Fig. \ref{fig:fid}. First, after discovering class-specific causal graphs, we rank the weights (and filters) in each sub-graph according to their highest effects (as described in eq. (\ref{eq:tce})). Then, we use these ranks to select the top-k critical neurons in each layer. As we observe in Fig. \ref{fig:fid}, the accuracy of all four models drastically drops after masking a small portion ($< 20\%$) of top critical neurons, and it is more evident on smaller architectures such as ResNet18 and MobileNetV2. In addition, these results describe another way of evaluating faithfulness since critical neurons encode the important features for predicting a specific class.
%
%Furthermore, we observe that the relation between performance drop and zeroing relevant neurons depends on the structural design of these models. Not necessarily the complexity related to the number of weights, which is $29$M in Tiny-ConvNeXt, $67$M in ResNet50V2, $11$M in ResNet18, and $2$M in MobileNetV2. In this analysis, we show another way to evaluate the trustworthiness (or faithfulness) of causal graphs to the model.    
%////////////////////////////////////////////////////////////////////////////////////////////////////////////////////////////////////////////////////////////////////////////////////////////////////////////////////////////
%//////////////////////////////////////////////////////////////////////////////////---APPLICATIONS---////////////////////////////////////////////////////////////////////////////////////////////
%%%%%%%%%%%%%%%%%%%%%%%%%%%%%%%%%%%%%%%%%%%%%%%%%%%%%%%%%%%%%%%%%%%%///////////////////////////////////////////////////////////////////////////////////////////////////////////////////////////////////////////////////////////////////////////////////////////////////////////////////
\section{Applications}
\begin{figure}
\centering
\footnotesize
\setlength{\tabcolsep}{1pt}
 \begin{tabular}{cc}
\includegraphics[width=0.5\linewidth]{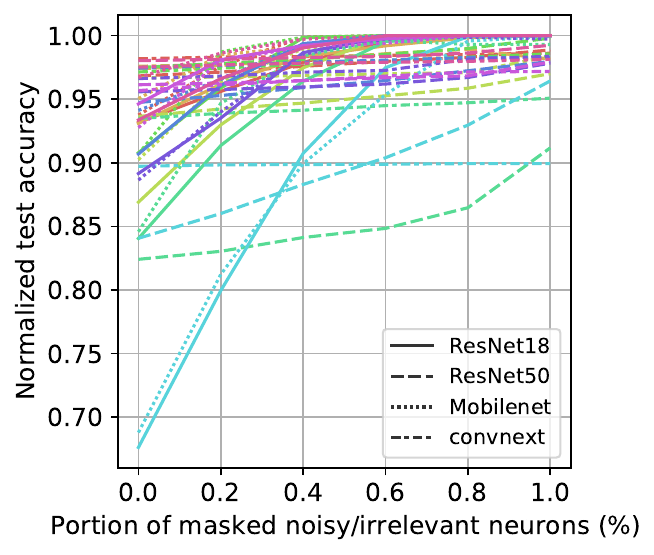} &
\includegraphics[width=0.5\linewidth]{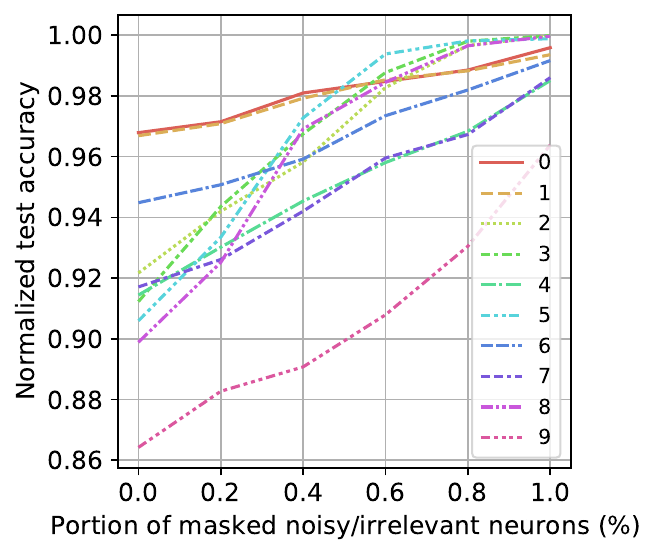} 
\end{tabular}
\caption{\textbf{Repair of model performance.} We show the test accuracy after masking out ($n\%$) of class-specific noisy filters in LeNet model for all categories of MNIST data (right), and in the 4 models for 10 representative (animal) categories of ImageNet data (left). Each color in left figure points to one different category and is fixed for each model.}
\label{fig:repair}
\end{figure}

\paragraph{Repairing model accuracy}
In many practical, real-world cases, we seek fast and effective ways to repair the model's behavior without requiring extensive retraining with large datasets. We can target the proposed explanation method to achieve this goal. Each causal explanatory graph measures the neurons' contributions to a specific class (or task) by intervening on the weights connecting the neurons to the class. More specifically, amortizing the strength of activation signals passing through particular paths that cause a drop in the model's performance or a wrong prediction. It is worth noting that this operation differs from model pruning since we only block these paths at inference time. Practically, we do this by masking out irrelevant weights (and filters for convolutional layers). The experiments show that our method can improve class prediction and correct wrong predictions. 
To illustrate these facts, we took the four models trained on ImageNet and considered 10 representative (animal) classes for evaluation in addition to the LeNet trained oon MNIST data. 
%and split the test sets between easy and hard samples. Easy samples are images that include relevant contexts of the object and refer to the correctly predicted inputs from each model. Hard samples are images missing some relevant contexts of the objects (e.g., cropped parts) or distracted by the presence of other labels making them misclassified. 
%
For each trained model, we select to mask out a portion of the irrelevant weights discovered by our method and evaluate how they perform on these samples. Figs \ref{fig:repair} shows test accuracy under a varying portion of the masked weights in all layers. 
%
%Fig \ref{fig:repair} shows the absolute number of masked weights (averaged over all classes) for each portion. 
% use the also qualitatively compared, for some models, between relevant features over easy and hard samples. We provide in Appendix using the attributions obtained by the causal graphs to visualize the locations of relevant features. We provide results in Appendix. 
%
%
%The literature has proven that DNN models are vulnerable to adversarial attacks and may fail in generalizing to outside of training distributions. We show in experiments that the causal graphs, estimated from training data, are robust and can be trusted in interpreting the inner behavior and output of these models. We apply our method to analyze the inner weaknesses in one of the models used in experiments. We examine a special type of attacks corresponding to data perturbations (e.g., noise) and the ResNet18 model as a use case. For each class, we perform 10 tests in which we attack every input by adding a small amount of Gaussian noise ($\mu=0.1, \sigma=0.1$). Then, we estimate the underlying causal graphs and compute their sensitivity w.r.t the noisy inputs. We show results in Fig. \ref{fig:robust}.   
%/////////////////////////////////////////////////////////////////////////////////////---CONCLUSION---////////////////////////////////////////////////////////////////////////////////////////////
%%%%%%%%%%%%%%%%%%%%%%%%%%%%%%%%%%%%%%%%%%%%%%%%%%%%%%%%%%%%%%%%%%%%%%
\section{Conclusion and discussions}
\label{concl}
We have presented a novel method for interpreting neural network behavior based on causal inference. It estimates the causal explanatory graphs that disentangle relevant knowledge hidden in the internal structure of DNNs, which is congenital to their predictions. Our methodology tests the hypothesis that path interventions for a parent neuron connected with target neurons in the subsequent layer will significantly affect the model's output. As a case study, we applied our method to vision models for object classification. The responses of causal filters are used to compare our approach to attribution methods quantitatively. This work is not aimed at extracting high-level abstractions that are interpretable to humans, which might be considered a limitation of our method. However, we seek to understand the inner working of the model and therefore provide a valuable tool for model monitoring and repair. We show that our method can be used to improve and fix the model without retraining, which makes it worthwhile and practical for real-world cases where extensive training data are not accessible, or retraining is computationally expensive. In future work, we will consider investigating further applications of our method. For instance, class-specific important neurons can be used with regularization methods in continual and few-shot learning. Our method's computational cost is reasonable, as shown in the supplementary, which facilitates its integration into other processes. The critical limitations of neuron importance methods are their high computational costs and sensitivity to superior correlations between neurons \cite{ghorbani_neuron_2020}. Relying on causal inference and path interventions allows for mitigating these limitations and provides robust interpretations.
%////////////////////////////////////////////////////////////////////////////////////////////////////////////////////////////////////////////////////////////////////////////////////////////////////////////////////////////
%//////////////////////////////////////////////////////////////////////////////////////////////////---END---////////////////////////////////////////////////////////////////////////////////////////////////////////////
%%%%%%%%%%%%%%%%%%%%%%%%%%%%%%%%%%%%%%%%%%%%%%%%%%%%%%%%%%%%%%%%%%%%%%

{\small
\bibliographystyle{ieee_fullname}
\bibliography{references.bib}
}

\end{document}